\documentclass[11pt]{article}

\usepackage[preprint]{acl}

\usepackage{times}
\usepackage{latexsym}
\usepackage[most]{tcolorbox}
\usepackage{caption}
\usepackage[T1]{fontenc}

\usepackage[utf8]{inputenc}

\usepackage{microtype}

\usepackage{inconsolata}

\usepackage{graphicx}
\usepackage{booktabs}
\usepackage{multirow}
\usepackage{amsmath}
\usepackage{fvextra}
\usepackage{pgfplots}
\usepackage{enumitem}
\usepgfplotslibrary{groupplots}
\pgfplotsset{compat=1.18}
\usetikzlibrary{calc}

\title{Benchmarking the Benchmarks:\\ Evaluating Benchmarks for Conversational Agents}

\author{
Noam Koren\thanks{Equal contribution.} \quad
Roy Bar-Haim\footnotemark[1] \quad
Abigail Goldsteen
\\
IBM Research
\\
\texttt{Noam.Koren1@ibm.com} \quad
\texttt{roybar@il.ibm.com} \quad
\texttt{abigailt@il.ibm.com}
 }

\definecolor{IssueGreen}{RGB}{0,255,0}
\definecolor{IssueYellow}{RGB}{220,190,0}
\definecolor{IssueCyan}{RGB}{0,0,255}
\definecolor{IssueMagenta}{RGB}{255,0,255}
\definecolor{IssueRed}{RGB}{255,0,0}

\begin{document}
\maketitle
\begin{abstract}
Task-oriented conversational agents are evaluated using curated or automatically generated benchmarks, yet benchmark quality is rarely assessed. Poor benchmarks may contain inconsistent tasks, simplistic scenarios, or limited policy coverage, leading to unreliable evaluations. We introduce a reference-free framework that uses LLM judges to assess benchmark consistency, complexity, and policy coverage, while providing actionable diagnostics of weaknesses. We validate the framework by demonstrating agreement with independent human annotations and by evaluating benchmarks generated by LLMs of varying capabilities, as well as benchmarks subjected to controlled quality-degrading perturbations. Across domains and judge models, the proposed metrics consistently distinguish between benchmark quality levels. We further demonstrate the framework’s applicability to manually curated benchmarks. Our framework offers a practical approach for evaluating synthetic and manually curated conversational-agent benchmarks.

\end{abstract}
\section{Introduction}

Task-oriented conversational agents are increasingly deployed in customer support, travel, and enterprise workflows \cite{kdd2025survey}. Unlike static question-answering systems, they must interact over multiple turns, follow domain-specific policies, operate in real-world environments, and use external tools. As their autonomy and complexity grow, reliable evaluation becomes increasingly important and challenging \cite{yehudai2026surveyevaluationllmbasedagents}.

A common approach is to evaluate agents using curated benchmarks with realistic tasks, user requests, environment states, and expected outcomes \citep{yao2024tau, barres2025tau2}. However, manual construction is costly, requiring domain expertise, scenario design, tool and database simulation, and specification of expected behavior. This has motivated growing interest in automatically generated benchmarks \citep{li2023api,qin2024toolllm,intellagent}, which forgo manual curation and its implicit quality control.

This shift raises a fundamental question: how should benchmark quality itself be evaluated? Synthetic benchmarks may contain inconsistent tasks, simplistic scenarios, or limited coverage of the capabilities and constraints they aim to test. Prior work shows that synthetic evaluation data can differ from human-curated data in representativeness, difficulty, and induced model rankings \citep{efficacy_synth_benchmark}.

Benchmark quality issues are not limited to synthetic data: a review of 445 benchmarks found pervasive construct-validity problems even in widely used, manually curated benchmarks \citep{bean2025construct-validity}, and $\tau$-bench, the basis for the benchmarks studied in this work, has also been shown to contain flawed and ambiguous tasks \citep{cuadron2025sabersmallactionsbig}.\footnote{$\tau^3$-bench, which we use for evaluation in \S\ref{sec:manual_benchmarks}, fixes issues found in earlier versions of $\tau$-bench.}\footnote{Sierra Blog also documents specific task fixes: \url{https://taubench.com/blog/tau3-task-fixes.html}.} Although recent studies have begun to assess benchmark quality more systematically \citep{qian2026benchmark2, iskander2024quality}, policy-grounded evaluation of conversational-agent benchmarks remains underexplored. Undetected benchmark flaws may therefore lead to unreliable conclusions about agent capabilities.

We address this gap by proposing reference-free metrics for evaluating conversational benchmarks with LLM judges \citep{zheng_LLMAJ}. The metrics capture multiple quality dimensions, including consistency, complexity, and coverage, without requiring comparison to existing datasets. They assess both task coherence and alignment with the domain policy that defines the intended agent behavior. Beyond aggregate scores, they identify concrete issues, such as inconsistent tasks and gaps in policy coverage, providing actionable diagnostics for benchmark developers. We demonstrate empirically that our metrics provide valuable insights across both synthetic and manually-created benchmarks.

A key challenge is validating that the metrics reflect meaningful quality differences. We construct benchmark suites with expected quality variation using two
complementary approaches: generating benchmarks with LLMs of different
capabilities, used as a proxy for expected quality, through the \textsc{IntellAgent} pipeline
\citep{intellagent}, and applying controlled perturbations that deliberately degrade benchmark quality. Across both settings, the metrics consistently distinguish quality levels across domains and LLM judges. We further validate the LLM-judge scores against independent human annotations, demonstrating medium-to-strong, statistically significant correlation between the two.

Our main contributions are:
\begin{enumerate}[nosep]
\item We identify and study the problem of evaluating conversational-agent benchmarks.
\item We propose multidimensional metrics that assess consistency, complexity, and coverage with respect to the domain policy.
\item We validate the metrics using benchmarks generated by different LLMs, controlled quality-degrading perturbations, and human evaluations.
\item We demonstrate the metrics on a manually curated benchmark ($\tau^3$-bench).
\end{enumerate}
\section{Benchmarks for Task-Oriented Conversational Agents}
\label{sec:background}

Recent task-oriented conversational-agent benchmarks, including \textsc{$\tau$-Bench}, \textsc{$\tau^2$-Bench}, and \textsc{IntellAgent} \citep{yao2024tau, barres2025tau2, intellagent}, evaluate agents through dynamic interactions with an LLM-based user simulator. The agent must fulfill user requests by combining natural-language dialogue with tool calls that access or update organizational information, while adhering to domain-specific policies. Our work assumes this benchmark structure and uses \textsc{IntellAgent} for benchmark generation.

A benchmark typically focuses on a specific \emph{domain}, such as \emph{airline} or \emph{retail}, which defines the \emph{environment} in which the agent operates. This environment comprises a set of available \emph{tools} and an associated \emph{database} containing relevant information, such as users, products, or orders. Each domain also includes a \emph{policy} that specifies the rules, constraints, and procedures governing agent behavior. The benchmark is designed to assess and evaluate the behaviors required by this policy.

Within each domain, a benchmark contains a collection of test cases, referred to as \emph{tasks}. Each task provides the information required to simulate the interaction and evaluate the agent trajectory. This includes a description of the user scenario, an initial database state, and expected agent behavior, including adherence to the domain policy, and, when available, a sequence of tool calls.

\section{Benchmark Metrics}
This work focuses on evaluating the conversational benchmarks used to test agents, rather than the agents themselves. While \emph{agent evaluation} measures an agent’s performance on a given benchmark, \emph{benchmark evaluation} assesses whether the benchmark is coherent, challenging, and representative of the behaviors defined by its domain policy.

\begin{figure*}[t]
\begin{center}
\includegraphics[width=0.9\textwidth]{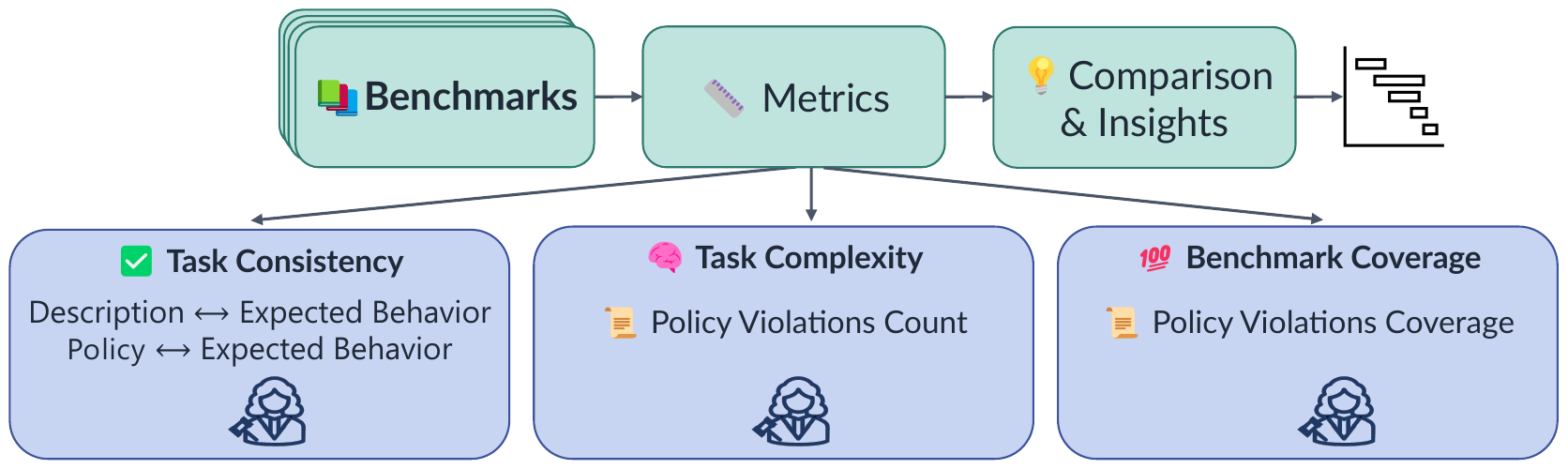}
\end{center}
\caption{Benchmark evaluation process
}
\label{fig:benchmark_evaluation}
\end{figure*}

We introduce four LLM-judge-based metrics that evaluate benchmark \emph{consistency} (\S\ref{ssec:consistency_metrics}), \emph{complexity}, and \emph{coverage} (\S\ref{ssec:complexity_and_coverage_metrics}). Together, these metrics provide a comprehensive view of benchmark quality and difficulty. Figure~\ref{fig:benchmark_evaluation} provides an overview of the benchmark evaluation process.

\paragraph{Notation.}
Let \(B = \left(P,{t_1,\dots,t_n}\right) \)
denote a benchmark consisting of a policy document $P$ and a set of $n$ tasks. Each task $t \in B$ represents a user scenario that the agent must handle in accordance with $P$ and is defined as
\begin{equation}
t = (d_t,e_t,i_t),
\label{eq:task}
\end{equation}
where \(d_t\) is the task description, including the user request; \(e_t\) specifies the expected agent behavior and should be consistent with the policy \(P\); and \(i_t\) is the initial database state. For any metric \(m\), let \(S_m(B)\) denote the score assigned to benchmark \(B\).

Figure~\ref{fig:task_example} illustrates a task and its components: the description \(d_t\), expected behavior \(e_t\), and initial state \(i_t\), which is omitted for brevity.

\begin{figure}[t]
    \centering
\fbox{
\begin{minipage}{0.95\columnwidth}
\textbf{Task \#0}

\textbf{Description:} User asks to modify reservation \texttt{RES12345}, asks about adding travel insurance, and later decides to cancel the reservation and ask about the refund.

\textbf{Expected Behavior:} The chatbot should request the user ID, explain that travel insurance cannot be added during modification, ask for the cancellation reason, and state that the refund will be sent to the original payment method within 5--7 business days.

\textbf{Initial state:} \texttt{<...>}
\end{minipage}
}
\caption{
Benchmark task \(t=(d_t,e_t,i_t)\), showing its description, expected behavior, and initial state.}
\label{fig:task_example}
\end{figure}

\subsection{Consistency Metrics}
\label{ssec:consistency_metrics}
Consistency measures whether the task description, domain policy, and expected behavior are aligned and free of contradictions. This helps identify and filter low-quality, ambiguous, or inconsistent tasks.

We define consistency through LLM-judged alignment between task components. Let \(J_m(\cdot)\) denote the LLM judge for metric \(m\). Given the task components, the judge assigns a score from \(1\) to \(10\), with higher scores indicating stronger alignment.

\paragraph{Description--Expected Behavior Alignment.}
This metric measures whether the expected behavior is consistent with the task description and adequately addresses the user’s intended goals.

\begin{equation}
M_{\mathrm{desc}}(t) = J_{\mathrm{desc}}(d_t, e_t).
\end{equation}

\paragraph{Policy--Expected Behavior Alignment.}
This metric measures whether the expected behavior complies with the domain policy for the described scenario. It penalizes unsupported actions and omissions of policy-required steps. Formally,
\begin{equation}
M_{\mathrm{pol}}(t) = J_{\mathrm{pol}}(d_t,e_t,P).
\end{equation}

For each $m \in {\mathrm{pol},\mathrm{desc}}$, the benchmark-level score is
\begin{equation}
S_m(B) = \frac{1}{n}\sum_{t \in B} M_m(t).
\end{equation}
Higher scores indicate greater consistency.

The LLM-judge prompts are provided in App.~\ref{app:judge_prompts}.
\subsection{Complexity and Coverage Metrics}
\label{ssec:complexity_and_coverage_metrics}

A key aspect of conversational-agent evaluation is testing whether the agent correctly identifies and handles requests that conflict with domain policies. For example, under the policy \emph{``The user may modify passenger details but not the number of passengers,''} a task may ask the agent to remove a passenger from an existing reservation. A high-quality benchmark should include broad and systematic coverage of such policy-violating scenarios.

A high-quality benchmark should also test compliant (``positive'') scenarios, not only violations. We focus on violations because they are more discriminative: a probe that only checks whether a policy item is \emph{relevant} to a scenario would yield near-complete coverage for almost any benchmark, since nearly every task engages most policy items, causing such a signal to saturate. Requiring a task to actively \emph{violate} a specific policy item is a much harder condition, and is therefore more informative.

Let \(L_P=(p_1,\ldots,p_N)\) denote a list of policy items, represented as annotated spans in the policy document \(P\). Let \(J_{\mathrm{violate}}(\cdot)\) be an LLM judge that identifies the policy items violated by a task based on its description and initial database state. We denote this set by

\begin{equation}
\begin{aligned}
V(t) &= J_{\mathrm{violate}}(P, L_P, t) \\
     &= \{ k : (d_t, i_t) \text{ violates } p_k \}.
\end{aligned}
\end{equation}
Let $T(p_k)$ be the set of tasks that violate policy item $p_k$:
\begin{equation}
\begin{aligned}
T(p_k) = \{t:k \in {V(t)}\}
\end{aligned}
\end{equation}

We define two benchmark-level metrics. The \textbf{Policy Violations per Task} score measures task complexity as the average number of policy violations identified per task:
\begin{equation}
S_{\mathrm{v\_count}}(B)
=\frac{1}{|B|}\sum_{t\in B}|V(t)|.
\end{equation}

The \textbf{Policy Violations Coverage} score measures the fraction of policy items violated by at least \(K\) tasks:
\begin{equation}
S_{\mathrm{v\_cov}}(B)=\frac{1}{N}|\{k: |T(p_k)| \ge K\}|
\end{equation}
We set \(K=3\) in our experiments.

The policy items were extracted semi-automatically. An initial set of policy spans was generated using \textsc{ToolGuard} \citep{zwerdling-etal-2025-towards} and subsequently reviewed and revised by one of the authors. The LLM judge prompt is provided in App.~\ref{app:judge_violations_prompt}.

\section{Metric Evaluation Benchmarks}
\label{sec:metric_evaluation}
To assess how well the proposed metrics distinguish quality differences among benchmarks, we construct a sequence of benchmarks of varying quality within the same domain -- a nontrivial challenge in itself.

We consider two approaches for generating such test benchmarks. First, we generate multiple benchmarks using a synthetic benchmark-generation pipeline. In each case, we vary selected LLM components while keeping the others fixed, using LLMs that span a range of capabilities (\S\ref{ssec:synthetic}). Second, we apply controlled perturbations to a given benchmark to degrade its quality (\S\ref{ssec:perturbed}).

The two approaches are complementary: generating benchmarks with LLMs of varying capabilities yields more realistic benchmarks and finer-grained quality differences, while the perturbation-based approach provides greater control over the differences in quality between the generated datasets.

\subsection{Synthetic Benchmark Generation}
\label{ssec:synthetic}
\begin{figure*}[t]
    \centering
\includegraphics[width=0.75\textwidth]{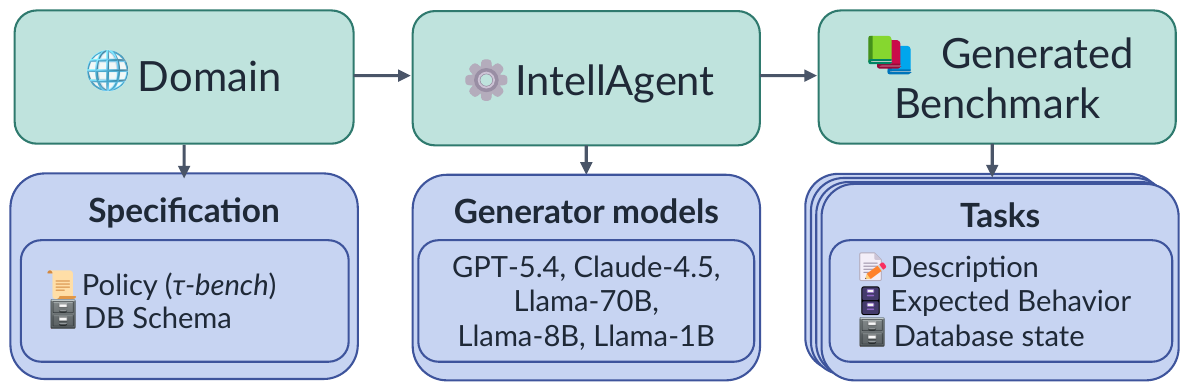}
    \caption{Benchmark generation setup 
    }
\label{fig:benchmark_generation}
\end{figure*}
We construct \emph{synthetic benchmarks} for each target domain using the \textsc{IntellAgent} framework \citep{intellagent}. Given a domain policy and database schema, \textsc{IntellAgent} first decomposes the policy into flows, such as \emph{booking a flight} or \emph{canceling a reservation}, and extracts the policies relevant to each flow. It then builds a policy graph whose nodes represent policy items and whose edges model their likelihood of co-occurring in the same task. For each task, a subset of policies is sampled through a random walk over the graph and used to generate the task description and expected behavior. Finally, the scenario is augmented with symbolic representations and instantiated entities, such as a \emph{user} or \emph{reservation}, and a corresponding initial database state is constructed.

By default, \textsc{IntellAgent} uses GPT-4o throughout the generation pipeline\footnote{Edge scoring uses GPT-4o-mini.}. To create benchmarks of varying quality, we replace GPT-4o in selected stages with LLMs spanning a broad range of capabilities. We use generator capability as a proxy for expected benchmark quality, a plausible assumption widely used in prior work: stronger models generally produce more
accurate and coherent outputs, and are often used to generate supervision for weaker models
\citep{grattafiori2024llama, 3692070.3692454, bubeck2023sparksartificialgeneralintelligence, kim-etal-2023-aligning}.

While weaker models can extract policies and generate task descriptions and expected behaviors, the final construction stages require a stronger model to complete reliably. We therefore retain GPT-4o for these stages in all configurations. Each generated event is converted into a task \(t\) with components \(d_t\), \(e_t\), and \(i_t\), as defined in Eq.~\ref{eq:task}. Figure~\ref{fig:benchmark_generation} illustrates the synthetic benchmark-generation process.

\subsection{Perturbed Benchmarks}
\label{ssec:perturbed}
Building on the synthetic benchmarks introduced in \S\ref{ssec:synthetic}, we construct \emph{perturbed benchmarks} by applying controlled perturbations to induce degradation in quality. We then evaluate whether the metric scores decrease in response. We apply the following perturbations to each synthetic benchmark.

\paragraph{Expected Behavior Swapping.}
To break the alignment between the task description and the expected behavior, we swap the expected behavior in $20\%, 40\%, 60\%,$ and $80\%$ of the tasks, while leaving the remainder of the benchmark unchanged. A useful metric should decrease as the perturbation rate increases.

\paragraph{Cross-Domain Policy Swapping.}
To evaluate policy-related metrics, we replace the policy of one domain with that of another domain and evaluate tasks under the mismatched policy. For example, airline tasks are evaluated against the retail policy, and vice versa. A useful policy metric should assign substantially lower scores under this perturbation.

\section{Metric Evaluation Experiments}
\label{sec:results}
\subsection{Experimental Setup}
\label{ssec:experimental_setup}

We use \textsc{IntellAgent} to generate five benchmarks for each of the two \textsc{$\tau$-Bench} domains, \emph{Airline} and \emph{Retail}. The benchmarks are generated using GPT-5.4, Claude-4.5-Sonnet, Llama-3.3-70B, Llama-3.1-8B, and Llama-3.2-1B. All configurations use the same generation pipeline (\S\ref{ssec:synthetic}) and differ only in the LLM used during the selected generation stages.

Each \emph{Airline} benchmark contains 100 tasks, and each \emph{Retail} benchmark contains 95, yielding 975 synthetic tasks in total. We evaluate the synthetic benchmarks using three LLM judges: GPT-5.4, Claude-4.5-Sonnet, and Gemini-2-Flash.

To quantify how well a metric induces a quality-based ordering over benchmarks, we define a meta-metric, the \emph{Benchmark Ordering Score}, applied to each of our four benchmark metrics. We partition the synthetic benchmarks in each domain into three tiers according to the capabilities of their generating LLMs:
\vspace{-0.01em}
\begin{itemize}[nosep, topsep=0pt, partopsep=0pt]
\item \emph{Top:} GPT-5.4 and Claude-4.5-Sonnet
\item \emph{Medium:} Llama-3.3-70B and Llama-3.1-8B
\item \emph{Small:} Llama-3.2-1B
\end{itemize}
\vspace{-0.01em}
We expect top-tier benchmarks to receive the highest metric scores, with medium-tier benchmarks scoring lower than top-tier ones but higher than small-tier benchmarks. This partial order induces a set of eight pairwise constraints that a metric score \(S(\cdot)\) should satisfy, e.g., \(S(B_{\text{GPT-5.4}}) > S(B_{\text{Llama-70B}})\), \(S(B_{\text{GPT-5.4}}) > S(B_{\text{Llama-1B}})\), and \(S(B_{\text{Llama-8B}}) > S(B_{\text{Llama-1B}})\). We define the Benchmark Ordering Score of a metric as the fraction of constraints satisfied. A score of \(1.0\) indicates that all pairwise order constraints are satisfied.

\subsection{Benchmark Ordering Across Generators}
\label{ssec:generator_ordering}
\begin{figure*}[ht!]
\centering
\begin{tikzpicture}

\pgfplotsset{
  sharedbarplot/.style={
    ybar,
    height=2.8cm,
    width=0.48\textwidth,
    bar width=5pt,
    enlarge x limits=0.25,
    symbolic x coords={GPT-5.4, Claude-Sonnet-4.5, Gemini-2-Flash},
    title style={font=\small},
    xtick=data,
    xticklabels={
      \textit{GPT-5.4},
      \textit{Claude-4.5},
      \textit{Gemini-2}
    },
    tick label style={font=\tiny},
    xlabel={Judge Model},
    ylabel={Score},
    label style={font=\small},
    ymin=0,
    ymajorgrids,
    major grid style={line width=.2pt, draw=gray!35},
    legend image code/.code={
      \draw[#1] (0cm,-0.1cm) rectangle (0.2cm,0.1cm);
    },
  }
}

\begin{groupplot}[
    group style={group size=2 by 4, horizontal sep=1.3cm, vertical sep=2.2cm},
    sharedbarplot
]

\nextgroupplot[
    title={\textbf{Airline}},
    legend to name=mainresultslegend,
    legend style={
  legend columns=-1,
  font=\small,
  draw=black,
  fill=white,
  /tikz/every even column/.append style={column sep=4pt}
}
]
\addplot[fill=purple] coordinates {(GPT-5.4, 6.27) (Claude-Sonnet-4.5, 8.04) (Gemini-2-Flash, 7.80)};
\addplot[fill=violet] coordinates {(GPT-5.4, 7.18) (Claude-Sonnet-4.5, 7.60) (Gemini-2-Flash, 8.85)};
\addplot[fill=cyan] coordinates {(GPT-5.4, 5.76) (Claude-Sonnet-4.5, 5.96) (Gemini-2-Flash, 7.64)};
\addplot[fill=teal] coordinates {(GPT-5.4, 4.14) (Claude-Sonnet-4.5, 4.52) (Gemini-2-Flash, 6.57)};
\addplot[fill=green!60!black] coordinates {(GPT-5.4, 4.05) (Claude-Sonnet-4.5, 4.06) (Gemini-2-Flash, 5.73)};
\legend{GPT-5.4, Claude-4.5, Llama-70B, Llama-8B, Llama-1B}

\nextgroupplot[title={\textbf{Retail}}]
\addplot[fill=purple] coordinates {(GPT-5.4,5.27) (Claude-Sonnet-4.5,6.86) (Gemini-2-Flash, 8.86)};
\addplot[fill=violet] coordinates {(GPT-5.4,7.83) (Claude-Sonnet-4.5,8.54) (Gemini-2-Flash, 7.18)};
\addplot[fill=cyan] coordinates {(GPT-5.4,7.05) (Claude-Sonnet-4.5,6.98) (Gemini-2-Flash, 8.58)};
\addplot[fill=teal] coordinates {(GPT-5.4,4.84) (Claude-Sonnet-4.5,5.60) (Gemini-2-Flash, 7.49)};
\addplot[fill=green!60!black] coordinates {(GPT-5.4,3.61) (Claude-Sonnet-4.5,4.39) (Gemini-2-Flash, 6.27)};

\nextgroupplot[title={\textbf{Airline}}]
\addplot[fill=purple] coordinates {(GPT-5.4, 6.64) (Claude-Sonnet-4.5, 7.48) (Gemini-2-Flash, 9.95)};
\addplot[fill=violet] coordinates {(GPT-5.4, 5.74) (Claude-Sonnet-4.5, 5.28) (Gemini-2-Flash, 9.80)};
\addplot[fill=cyan] coordinates {(GPT-5.4, 4.41) (Claude-Sonnet-4.5, 4.13) (Gemini-2-Flash, 8.75)};
\addplot[fill=teal] coordinates {(GPT-5.4, 3.47) (Claude-Sonnet-4.5, 3.75) (Gemini-2-Flash, 7.52)};
\addplot[fill=green!60!black] coordinates {(GPT-5.4, 3.06) (Claude-Sonnet-4.5, 3.13) (Gemini-2-Flash, 5.15)};

\nextgroupplot[title={\textbf{Retail}}]
\addplot[fill=purple] coordinates {(GPT-5.4,6.43) (Claude-Sonnet-4.5,7.89) (Gemini-2-Flash, 9.87)};
\addplot[fill=violet] coordinates {(GPT-5.4,5.49) (Claude-Sonnet-4.5,6.93) (Gemini-2-Flash, 9.38)};
\addplot[fill=cyan] coordinates {(GPT-5.4,4.58) (Claude-Sonnet-4.5,5.56) (Gemini-2-Flash, 9.23)};
\addplot[fill=teal] coordinates {(GPT-5.4,3.68) (Claude-Sonnet-4.5,4.32) (Gemini-2-Flash, 7.78)};
\addplot[fill=green!60!black] coordinates {(GPT-5.4,3.33) (Claude-Sonnet-4.5,4.37) (Gemini-2-Flash, 6.03)};
\nextgroupplot[title={\textbf{Airline}}]
\addplot[fill=purple] coordinates {(GPT-5.4, 2.28) (Claude-Sonnet-4.5, 1.97) (Gemini-2-Flash, 1.95)};
\addplot[fill=violet] coordinates {(GPT-5.4, 1.54) (Claude-Sonnet-4.5, 1.57) (Gemini-2-Flash, 1.74)};
\addplot[fill=cyan] coordinates {(GPT-5.4, 1.07) (Claude-Sonnet-4.5, 0.96) (Gemini-2-Flash, 1.03)};
\addplot[fill=teal] coordinates {(GPT-5.4, 1.12) (Claude-Sonnet-4.5, 1.08) (Gemini-2-Flash, 1.17)};
\addplot[fill=green!60!black] coordinates {(GPT-5.4, 0.53) (Claude-Sonnet-4.5, 0.53) (Gemini-2-Flash, 0.91)};

\nextgroupplot[title={\textbf{Retail}}]
\addplot[fill=purple] coordinates {(GPT-5.4, 2.18) (Claude-Sonnet-4.5, 1.43) (Gemini-2-Flash, 1.46)};
\addplot[fill=violet] coordinates {(GPT-5.4, 1.75) (Claude-Sonnet-4.5, 1.4) (Gemini-2-Flash, 1.56)};
\addplot[fill=cyan] coordinates {(GPT-5.4, 1.32) (Claude-Sonnet-4.5, 0.85) (Gemini-2-Flash, 1.21)};
\addplot[fill=teal] coordinates {(GPT-5.4, 1.23) (Claude-Sonnet-4.5, 0.81) (Gemini-2-Flash, 1.09)};
\addplot[fill=green!60!black] coordinates {(GPT-5.4, 0.07) (Claude-Sonnet-4.5, 0.03) (Gemini-2-Flash, 0.15)};

\nextgroupplot[title={\textbf{Airline}}]		
\addplot[fill=purple] coordinates {(GPT-5.4, 0.58) (Claude-Sonnet-4.5, 0.58) (Gemini-2-Flash, 0.5)};		
\addplot[fill=violet] coordinates {(GPT-5.4, 0.54) (Claude-Sonnet-4.5, 0.5) (Gemini-2-Flash, 0.54)};
\addplot[fill=cyan] coordinates {(GPT-5.4, 0.5) (Claude-Sonnet-4.5, 0.38) (Gemini-2-Flash, 0.38)};
\addplot[fill=teal] coordinates {(GPT-5.4, 0.35) (Claude-Sonnet-4.5, 0.38) (Gemini-2-Flash, 0.42)};	
\addplot[fill=green!60!black] coordinates {(GPT-5.4, 0.23) (Claude-Sonnet-4.5, 0.23) (Gemini-2-Flash, 0.27)};

\nextgroupplot[title={\textbf{Retail}}]
\addplot[fill=purple] coordinates {(GPT-5.4, 0.72) (Claude-Sonnet-4.5, 0.52) (Gemini-2-Flash, 0.4)};
\addplot[fill=violet] coordinates {(GPT-5.4, 0.6) (Claude-Sonnet-4.5, 0.6) (Gemini-2-Flash, 0.56)};
\addplot[fill=cyan] coordinates {(GPT-5.4, 0.64) (Claude-Sonnet-4.5, 0.4) (Gemini-2-Flash, 0.44)};
\addplot[fill=teal] coordinates {(GPT-5.4, 0.56) (Claude-Sonnet-4.5, 0.44) (Gemini-2-Flash, 0.52)};
\addplot[fill=green!60!black] coordinates {(GPT-5.4, 0.04) (Claude-Sonnet-4.5, 0) (Gemini-2-Flash, 0.08)};

\end{groupplot}

\node[font=\small\bfseries] at ($(group c1r1.north)!0.5!(group c2r1.north)+(0,0.75cm)$)
{Description--Expected Behavior Alignment};

\node[font=\small\bfseries] at ($(group c1r2.north)!0.5!(group c2r2.north)+(0,0.75cm)$)
{Policy--Expected Behavior Alignment};

\node[font=\small\bfseries] at ($(group c1r3.north)!0.5!(group c2r3.north)+(0,0.75cm)$)
{Policy Violations Count};

\node[font=\small\bfseries] at ($(group c1r4.north)!0.5!(group c2r4.north)+(0,0.75cm)$)
{Policy Violations Coverage};

\node[font=\small\bfseries]
at ($(group c1r4.south)!0.5!(group c2r4.south)+(0,-1.0cm)$)
{Benchmark Generator};

\node at ($(group c1r4.south)!0.5!(group c2r4.south)+(0,-1.7cm)$)
{\pgfplotslegendfromname{mainresultslegend}};
\end{tikzpicture}

\caption{Benchmark quality metrics across domains, generated benchmarks, and judge models. Colors denote the benchmark generator model, while groups correspond to the judge model. We report Description--Expected Behavior Alignment, Policy--Expected Behavior Alignment, and Policy Violations for the Airline and Retail domains, as judged by GPT-5.4, Claude-Sonnet-4.5, and Gemini-2-Flash. 
}
\label{fig:results}
\end{figure*}
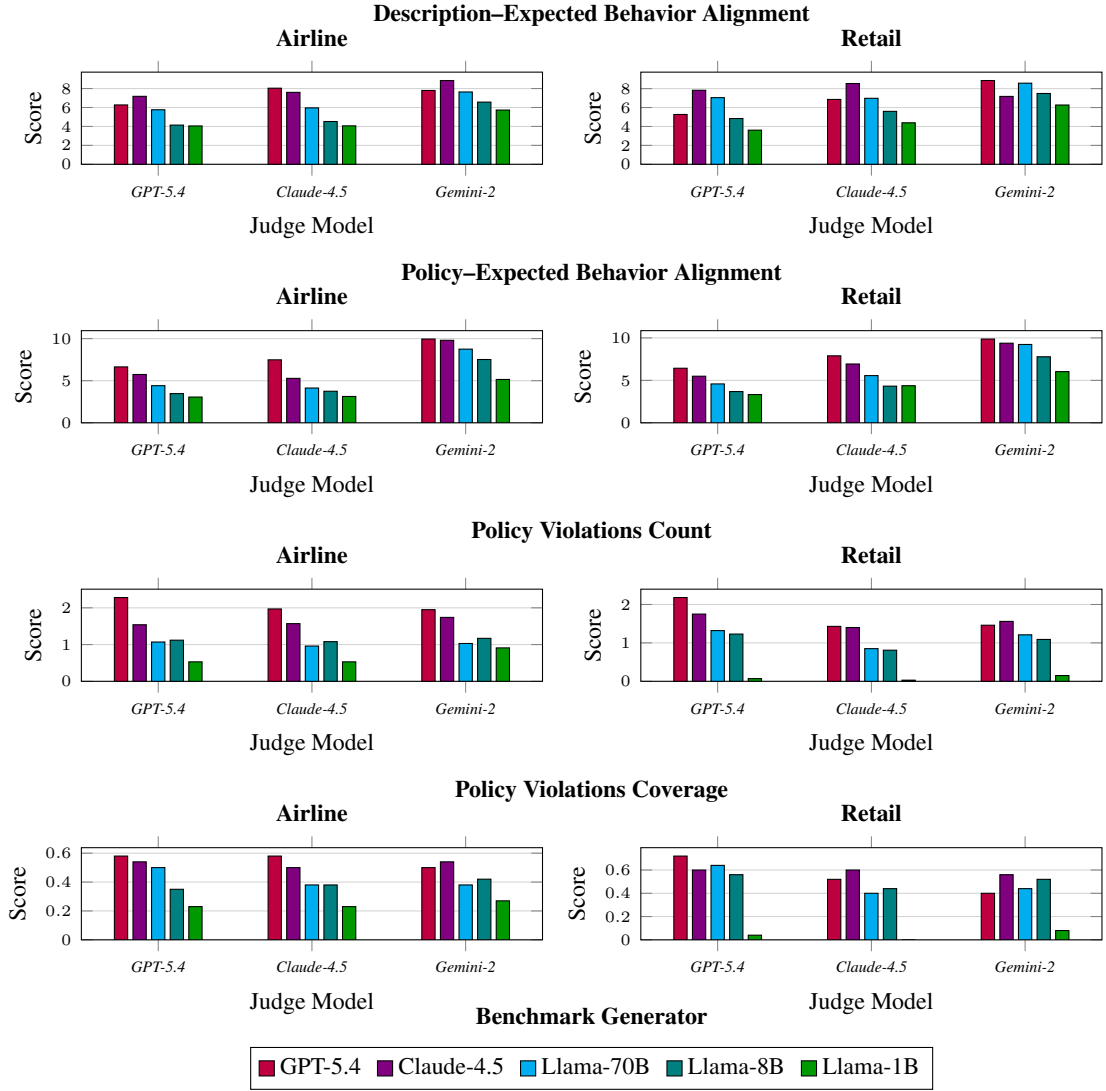

Figure~\ref{fig:results} shows substantial score gaps between
benchmarks, with metrics consistently assigning higher scores to
benchmarks generated by stronger models (e.g., GPT-5.4 and
Claude-4.5-Sonnet), across metrics, domains, and judges. The Benchmark
Ordering Scores, summarized in Table~\ref{tab:metric_ordering_score}
(App.~\ref{app:ordering_scores}), quantify this observation: all
metrics achieve perfect ordering (1.0) in the Airline domain, and a
mean of 0.92 in Retail. Policy Violations Count achieves perfect
ordering across all judges and domains, and the overall mean ordering
score is slightly higher for GPT-5.4 and Claude-Sonnet-4.5 than for
Gemini-2-Flash (0.97 vs.\ 0.94).

These results provide strong evidence that our benchmark-quality
metrics reliably capture quality differences among synthetic
benchmarks.

\subsection{Detecting Controlled Quality Degradation}
\label{ssec:perturbation_results}

We next evaluate the sensitivity of the proposed metrics to the controlled
perturbations defined in \S\ref{ssec:perturbed}, using Gemini-2-Flash as
the LLM judge.

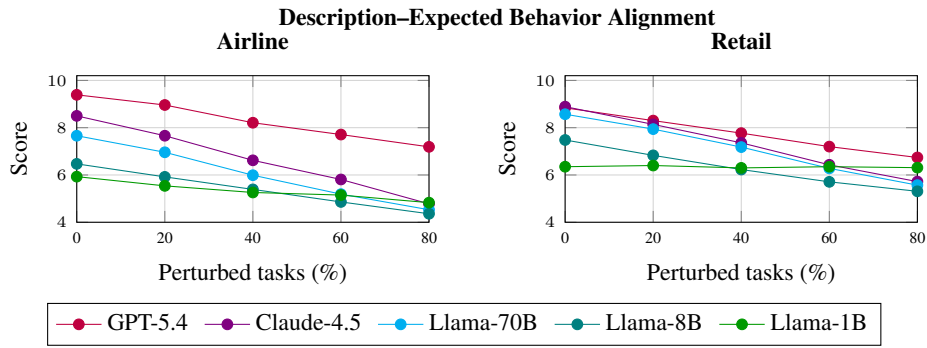
\begin{figure*}[ht!]
\centering
\begin{tikzpicture}
\begin{groupplot}[
    group style={group size=2 by 1, horizontal sep=1.8cm},
    width=0.39\textwidth,
    height=0.22\textwidth,
    xmin=0, xmax=80,
    xtick={0,20,40,60,80},
    xticklabels={0,20,40,60,80},
    xlabel={Perturbed tasks (\%)},
    ylabel={Score},
    grid=both,
    major grid style={line width=.2pt, draw=gray!35},
    minor grid style={line width=.1pt, draw=gray!20},
    tick label style={font=\tiny},
    label style={font=\small},
    title style={font=\small},
    legend style={font=\small, draw=black, fill=white},
    legend cell align=left
]

\nextgroupplot[
    title={\textbf{Airline}},
    ymin=4.0, ymax=10.2,
    legend style={
      at={(1.1,-0.55)},
  anchor=north,
  legend columns=-1,
  font=\small,
  draw=black,
  fill=white,
  /tikz/every even column/.append style={column sep=4pt}
  }
]
\addplot[color=purple, mark=*] coordinates {(0,9.39) (20,8.96) (40,8.21) (60,7.71) (80,7.19)};
\addplot[color=violet, mark=*] coordinates {(0,8.50) (20,7.66) (40,6.62) (60,5.81) (80,4.78)};
\addplot[color=cyan, mark=*] coordinates {(0,7.66) (20,6.96) (40,5.99) (60,5.19) (80,4.53)};
\addplot[color=teal, mark=*] coordinates {(0,6.47) (20,5.92) (40,5.39) (60,4.86) (80,4.36)};
\addplot[color=green!60!black, mark=*] coordinates {(0,5.93) (20,5.54) (40,5.26) (60,5.15) (80,4.83)};
\legend{GPT-5.4, Claude-4.5, Llama-70B, Llama-8B, Llama-1B}
\nextgroupplot[
    title={\textbf{Retail}},
    ymin=4.0, ymax=10.2,
    legend to name=perturbabslegend_retail,
    legend columns=-1,
]
\addplot[color=purple, mark=*] coordinates {(0,8.84) (20,8.30) (40,7.77) (60,7.20) (80,6.74)};
\addplot[color=violet, mark=*] coordinates {(0,8.89) (20,8.14) (40,7.36) (60,6.43) (80,5.72)};
\addplot[color=cyan, mark=*] coordinates {(0,8.57) (20,7.94) (40,7.18) (60,6.28) (80,5.57)};
\addplot[color=teal, mark=*] coordinates {(0,7.48) (20,6.83) (40,6.23) (60,5.71) (80,5.31)};
\addplot[color=green!60!black, mark=*] coordinates {(0,6.35) (20,6.40) (40,6.30) (60,6.35) (80,6.31)};

\end{groupplot}
\node[font=\small\bfseries] at ($(group c1r1.north)!0.5!(group c2r1.north)+(0,0.75cm)$)
{Description--Expected Behavior Alignment};
\end{tikzpicture}
\caption{Mean Description–Expected Behavior Alignment-metric scores assigned by Gemini under increasing perturbation rates, where expected behaviors are swapped in 0\%, 20\%, 40\%, 60\%, and 80\% of the tasks.}
\label{fig:perturbation_absolute_gemini}

\end{figure*}

\paragraph{Expected Behavior Swapping.}
We apply this perturbation to the
\emph{Description--Expected Behavior Alignment} metric, which measures
task-level consistency. Figure~\ref{fig:perturbation_absolute_gemini}
shows that the metric score decreases consistently as the proportion of
tasks with swapped expected behaviors increases. This pattern holds
across both domains and all benchmark generators, demonstrating that the
metric is sensitive to controlled degradation in task coherence.

\begin{figure*}[t]
\centering

\newcommand{\panelw}{0.305\textwidth}
\newcommand{\groupsep}{0.45cm}
\newcommand{\axw}{0.325\linewidth}
\newcommand{\axh}{0.50\linewidth}

\makebox[\textwidth][c]{

\begin{minipage}[t]{\panelw}
\centering
\begin{tikzpicture}
\begin{groupplot}[
  group style={group size=2 by 1, horizontal sep=\groupsep},
  scale only axis,
  width=\axw, height=\axh,
  xmin=0, xmax=1,
  xtick={0,1},
  xticklabels={{\makebox[0pt][l]{Orig.}},{{\makebox[0pt][r]{Swapped}}}},
  xticklabel style={align=center, font=\scriptsize},
  tick label style={font=\scriptsize},
  label style={font=\small},
  title style={font=\small},
  grid=both,
  major grid style={line width=.2pt, draw=gray!35},
  minor grid style={line width=.1pt, draw=gray!20},
  legend cell align=left,
]

\nextgroupplot[
  title={\textbf{Airline}},
  ymin=3.0, ymax=10.2,
  ylabel={Score},
  legend to name=sharedlegend,
  legend columns=-1,
  legend style={
    font=\footnotesize,
    draw=black,
    fill=white,
    /tikz/every even column/.append style={column sep=6pt}
  }
]
\addplot[color=purple, mark=*] coordinates {(0,9.95) (1,4.56)};
\addlegendentry{GPT-5.4}
\addplot[color=violet, mark=*] coordinates {(9.80,0) (1,4.58)};
\addlegendentry{Claude-4.5}
\addplot[color=cyan, mark=*]   coordinates {(0,8.75) (1,4.64)};
\addlegendentry{Llama-70B}
\addplot[color=teal, mark=*]   coordinates {(0,7.52) (1,4.39)};
\addlegendentry{Llama-8B}
\addplot[color=green!60!black, mark=*] coordinates {(0,5.15) (1,4.29)};
\addlegendentry{Llama-1B}

\nextgroupplot[
  title={\textbf{Retail}},
  ymin=3.0, ymax=10.2,
  ylabel={Score},
  ylabel style={opacity=0},
  yticklabels={}
]
\addplot[color=purple, mark=*] coordinates {(0,9.87) (1,3.42)}; 
\addplot[color=violet, mark=*] coordinates {(0,9.38) (1,4.25)}; 
\addplot[color=cyan, mark=*]   coordinates {(0,9.23) (1,5.36)}; 
\addplot[color=teal, mark=*]   coordinates {(0,7.78) (1,4.16)}; 
\addplot[color=green!60!black, mark=*] coordinates {(0,6.03) (1,3.27)}; 

\end{groupplot}

\node[font=\small\bfseries] at
($(group c1r1.north)!0.5!(group c2r1.north)+(0,0.85cm)$) {Policy–Expected Behavior Alignment};

\end{tikzpicture}
\end{minipage}
\hspace{\groupsep}

\begin{minipage}[t]{\panelw}
\centering
\begin{tikzpicture}
\begin{groupplot}[
  group style={group size=2 by 1, horizontal sep=\groupsep},
  scale only axis,
  width=\axw, height=\axh,
  xmin=0, xmax=1,
  xtick={0,1},
  xticklabels={{\makebox[0pt][l]{Orig.}},{{\makebox[0pt][r]{Swapped}}}},
  xticklabel style={align=center, font=\scriptsize},
  tick label style={font=\scriptsize},
  label style={font=\small},
  title style={font=\small},
  grid=both,
  major grid style={line width=.2pt, draw=gray!35},
  minor grid style={line width=.1pt, draw=gray!20},
]

\nextgroupplot[title={\textbf{Airline}}, ymin=0.0, ymax=2.1, ylabel={Score}]
\addplot[color=purple, mark=*] coordinates {(0,1.95) (1,1.11) };
\addplot[color=violet, mark=*] coordinates {(0,1.74) (1,1.11)};
\addplot[color=cyan, mark=*]   coordinates {(0,1.03) (1,0.62)};
\addplot[color=teal, mark=*]   coordinates {(0,1.17) (1,0.71)};
\addplot[color=green!60!black, mark=*] coordinates {(0,0.91) (1,0.52) };

\nextgroupplot[
  title={\textbf{Retail}},
  ymin=0.0, ymax=2.1,
  ylabel={Score},
  ylabel style={opacity=0},
  yticklabels={}
]
\addplot[color=purple, mark=*] coordinates {(0,1.46) (1,0.72)}; 
\addplot[color=violet, mark=*] coordinates {(0,1.56) (1,0.69)}; 
\addplot[color=cyan, mark=*]   coordinates {(0,1.21) (1,0.51)}; 
\addplot[color=teal, mark=*]   coordinates {(0,1.09) (1,0.43)}; 
\addplot[color=green!60!black, mark=*] coordinates {(0,0.15) (1,0.02)}; 

\end{groupplot}

\node[font=\small\bfseries] at
($(group c1r1.north)!0.5!(group c2r1.north)+(0,0.85cm)$) {Policy Violations Count};

\end{tikzpicture}
\end{minipage}
\hspace{\groupsep}

\begin{minipage}[t]{\panelw}
\centering
\begin{tikzpicture}
\begin{groupplot}[
  group style={group size=2 by 1, horizontal sep=\groupsep},
  scale only axis,
  width=\axw, height=\axh,
  xmin=0, xmax=1,
  xtick={0,1},
  xticklabels={{\makebox[0pt][l]{Orig.}},{{\makebox[0pt][r]{Swapped}}}},
  xticklabel style={align=center, font=\scriptsize},
  tick label style={font=\scriptsize},
  label style={font=\small},
  title style={font=\small},
  grid=both,
  major grid style={line width=.2pt, draw=gray!35},
  minor grid style={line width=.1pt, draw=gray!20},
]

\nextgroupplot[title={\textbf{Airline}}, ymin=0.0, ymax=0.6, ylabel={Score}]
\addplot[color=purple, mark=*] coordinates {(0,0.5) (1,0.32)};
\addplot[color=violet, mark=*] coordinates {(0,0.54) (1,0.24)};
\addplot[color=cyan, mark=*]   coordinates {(0,0.38) (1,0.16)};
\addplot[color=teal, mark=*]   coordinates {(0,0.42) (1,0.16)};
\addplot[color=green!60!black, mark=*] coordinates {(0,0.27) (1,0.08)};

\nextgroupplot[
  title={\textbf{Retail}},
  ymin=0.0, ymax=0.6,
  ylabel={Score},
  ylabel style={opacity=0},
  yticklabels={}
]
\addplot[color=purple, mark=*] coordinates {(0,0.40) (1,0.15)}; 
\addplot[color=violet, mark=*] coordinates {(0,0.56) (1,0.15)}; 
\addplot[color=cyan, mark=*]   coordinates {(0,0.44) (1,0.15)}; 
\addplot[color=teal, mark=*]   coordinates {(0,0.52) (1,0.15)}; 
\addplot[color=green!60!black, mark=*] coordinates {(0,0.08) (1,0)}; 

\end{groupplot}

\node[font=\small\bfseries] at
($(group c1r1.north)!0.5!(group c2r1.north)+(0,0.85cm)$) {Policy Violations Coverage};

\end{tikzpicture}
\end{minipage}

}

\vspace{2mm}
\begin{center}
\begin{tikzpicture}
\node {\pgfplotslegendfromname{sharedlegend}};
\end{tikzpicture}
\end{center}

\caption{Cross-domain perturbation effects on metric scores, comparing original benchmarks to a “corrupted” setting in which airline and retail policies are swapped. Metric scores are computed using \emph{Gemini-2} as the judge.}
\label{fig:perturbation_absolute_gemini_3groups}
\end{figure*}
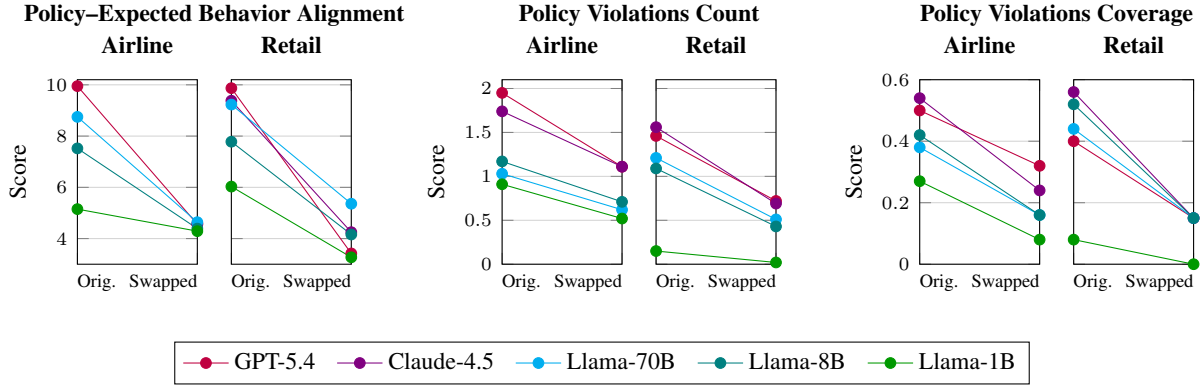

\paragraph{Cross-Domain Policy Swapping.}
We apply this perturbation to the three policy-grounded metrics:
\emph{Policy--Expected Behavior Alignment},
\emph{Policy Violations Count}, and
\emph{Policy Violations Coverage}. As shown in
Figure~\ref{fig:perturbation_absolute_gemini_3groups}, all three metrics
decrease substantially when tasks are evaluated against a policy from
the other domain. The consistent decline across domains and benchmark
generators shows that these metrics detect mismatches between benchmark
tasks and their associated domain policy.

\subsection{Human Validation}
\label{sec:human-validation}

As an independent validation of the LLM-based metrics, we conducted a human evaluation study. A human annotator rated 50 tasks, sampled across benchmarks generated by all five models, along the three task-level measures in our framework: Description--Expected Behavior Alignment, Policy--Expected Behavior Alignment, and Policy Violations Count. We computed Kendall's
\(\tau_b\) correlation \citep{kendall1945treatment} between the human ratings and the scores assigned by each LLM judge, pooling all 50 tasks into a single sample.

As shown in Table~\ref{tab:human-validation} (App.~\ref{app:human-validation}), all nine correlations (3 metrics \(\times\) 3 LLM judges) are positive and statistically significant. The
\(\tau_b\) values indicate moderate-to-strong agreement, ranging from \(0.32\) to \(0.67\), with raw
\(p\)-values ranging from \(<10^{-6}\) to \(0.011\). Agreement is
strongest for Policy Violations Count, with correlations ranging from
\(0.55\) to \(0.67\). Overall, these results provide independent
evidence that the LLM judges produce task-level rankings that align
with human assessments. Together with the benchmark ordering results
in \S\ref{ssec:generator_ordering}, this human validation supports our
assumption in \S\ref{ssec:synthetic} that more capable models tend to
generate benchmarks of higher quality and complexity. Additional
details are provided in App.~\ref{app:human-validation}.

\subsection{Qualitative Analysis}
\label{sec:example_issues}

Following the quantitative validation of our metrics, we examine their
value as diagnostic tools, using \emph{Policy--Expected Behavior
Alignment} as a case study. We analyze low-scoring \emph{Airline} tasks
and the corresponding rationales produced by the Gemini-2-Flash judge.
We manually review the examples, use GPT-5 to cluster recurring failure
patterns, and then manually verify and refine the resulting categories.

The analysis identifies five recurring issues in the generated expected
behaviors:
\begin{enumerate}[nosep]
    \item Hallucinated workflow stages or unsupported actions;
    \item Missing required information collection;
    \item Missing explicit confirmation before an action;
    \item Incorrect compensation handling; and
    \item Forbidden actions.
\end{enumerate}

These categories demonstrate how the metric can distinguish among
different benchmark weaknesses rather than providing only an aggregate
score. App.~\ref{app:policy_issue_examples} presents representative
examples of each issue type.

\section{Evaluating Manually-Constructed Benchmarks}
\label{sec:manual_benchmarks}

Beyond synthetic benchmarks, our metrics also apply to manually
constructed ones. We illustrate this on \textsc{\(\tau^3\)-Bench},
evaluating all 50 \emph{Airline} tasks\footnote{We focused on the
\emph{Airline} domain, as expected behavior annotations were missing
from most \emph{Retail} tasks.} with Claude-4.5-Sonnet and GPT-5.4 as
judges, using metric prompts minimally adapted for structural
differences from \textsc{IntellAgent} (App.~\ref{app:tau3_eval}).
Table~\ref{tab:tau} reports the results.

\begin{table}[t]
\centering
\small
\begin{tabular}{lrr}
\toprule
 & Claude-4.5-Sonnet & GPT-5.4 \\
\midrule
Desc.--Expected        & 3.50 & 4.54 \\
Policy--Expected       & 5.28 & 3.80 \\
Violation Coverage     & 0.31 & 0.46 \\
Violations Count    & 0.38 & 0.60 \\
\bottomrule
\end{tabular}
\caption{Metric scores on the manually curated
\textsc{\(\tau^3\)-Bench} benchmark (\emph{Airline} domain).}
\label{tab:tau}
\end{table}

The \(\tau^3\)-Bench scores broadly fall within the range observed for
the lower-scoring \textsc{IntellAgent} benchmarks generated by
Llama-3.1-8B and Llama-3.2-1B. For the consistency metrics, this should
be read in light of differences in expected-behavior granularity: in
\(\tau^3\)-Bench, the expected behavior typically specifies only the
final outcome (e.g., whether a cancellation request is approved),
whereas \textsc{IntellAgent} benchmarks generated by stronger models
describe the full policy-compliant workflow, including information
collection, confirmation steps, and applicable procedures. Since
Policy--Expected Behavior Alignment penalizes missing policy-required
steps, it scores concise outcome-level annotations lower even when they
are correct. \(\tau^3\)-Bench also shows substantially fewer policy
violations per task (0.38--0.60) than the strongest \textsc{IntellAgent}
benchmarks (1.5--2.3), indicating that top-generator \textsc{IntellAgent}
tasks are considerably more adversarial and policy-complex. Finally,
policy-violation coverage (0.31--0.46) leaves room for improvement,
indicating that a substantial share of policy items are not challenged
by any task in the benchmark.

\section{Related Work}
Existing LLM-agent benchmarks evaluate a broad range of capabilities
\cite{kdd2025survey, yehudai2026surveyevaluationllmbasedagents}, from
tool use \cite{berkeley-function-calling-leaderboard} and planning
\cite{valmeekam2023planbench} to specific settings such as web
interaction \cite{yao2022webshop, zhou2023webarena} and software
engineering \cite{Jimenez2023SWEbenchCL, miserendino2025swe}, and more
recently, general-purpose agents across multiple tasks and
environments \cite{trivedi-etal-2024-appworld, xu2024theagentcompany}.

Within this broader landscape, we focus on benchmarks for task-oriented
conversational agents, such as \textsc{\(\tau\)-Bench} and
\textsc{IntellAgent} \cite{yao2024tau, intellagent}, which evaluate
agents through dynamic interactions with an LLM-based user simulator
(\S\ref{sec:background}). Whereas \textsc{\(\tau\)-Bench} contains
manually constructed tasks, \textsc{IntellAgent} generates synthetic
tasks from a domain policy and database schema
(\S\ref{ssec:synthetic}).

Although these benchmarks support agent evaluation, their own quality has received comparatively little systematic attention. This issue is especially important for synthetic benchmarks, which often lack direct human validation. A common approach is to compare them with manually constructed references. For example, \citeauthor{intellagent} validate \textsc{IntellAgent} by measuring correlations between agent performance on its tasks and on \textsc{\(\tau\)-Bench}. Similarly, \citet{efficacy_synth_benchmark} compare synthetic and human-curated datasets using model-ranking agreement, fidelity measures, and statistical similarity. However, these approaches provide only indirect evidence of benchmark quality and require suitable reference datasets, which may not always be available. Outside the conversational-agent setting, \citet{qian2026benchmark2} assess closed-form LLM benchmarks (e.g., \textsc{MMLU}, \textsc{ARC}) using model-ranking consistency across benchmarks, discriminative power across models, and detection of anomalous within-family model comparisons; however, their approach requires evaluating many LLMs on each benchmark, which is considerably more costly for open-ended, task-oriented conversational settings than for closed-form benchmarks, and is not required by our metrics.

To the best of our knowledge, our work is the first to systematically evaluate benchmarks for conversational agents. Unlike previous benchmark-evaluation approaches, ours does not require a reference benchmark or running a large set of agents or LLMs on the benchmark being evaluated. Building on \textsc{\(\tau\)-Bench} and \textsc{IntellAgent}, our metrics quantify benchmark quality and complexity while also identifying concrete weaknesses, such as inconsistent tasks and insufficient policy coverage.

\section{Conclusion}
We introduced a framework for evaluating the quality of
conversational-agent benchmarks, defining metrics that capture
task-level consistency and complexity, and benchmark-level policy
coverage. Beyond aggregate scores, these metrics provide actionable
diagnostics, surfacing low-quality tasks and gaps in policy coverage.

We validated the metrics using synthetic benchmarks from LLMs of
varying capabilities, controlled perturbations, and an independent
human evaluation, and illustrated their use on a manually constructed
benchmark. The metrics reliably capture quality differences, remain
sensitive to controlled degradation, and align with human judgments.
Together, they promote more systematic, reliable evaluation of
conversational-agent benchmarks.
\section*{Limitations}

This work focuses on conversational-agent benchmarks following
\textsc{\(\tau\)-Bench}- and \textsc{IntellAgent}-style evaluation
frameworks. Accordingly, our metrics primarily assess task consistency,
complexity, and alignment with domain policies. 

Our synthetic-benchmark experiments rely on the policy-driven
\textsc{IntellAgent} generation pipeline. Other strategies, such as
tool-driven generation, may produce different task distributions.
Future work should therefore evaluate the proposed metrics across a
broader range of benchmark-generation methods and domains.

Finally, we do not consider non-conversational agent settings, such as
code generation or computer interaction, which involve different
interaction structures and evaluation challenges.
\bibliography{custom}
\appendix
\section{LLM-as-a-Judge Prompts for Consistency Metrics}
\label{app:judge_prompts}

To evaluate benchmark consistency, we use LLM-based judges that score the alignment between different task components. Each judge receives the relevant components of a task and produces a score on a $1$--$10$ scale, where higher scores indicate stronger alignment and consistency.

The judges are prompted with structured evaluation instructions and asked to return a numeric score reflecting the degree of alignment.

\subsection{Description--Expected Behavior Alignment}
\label{app:desc_expected}
This judge evaluates whether the expected behavior accurately reflects the task description and intended user goals.

\paragraph{Inputs}

\begin{itemize}
\item Task description $d_t$
\item Expected behavior $e_t$
\end{itemize}

\paragraph{Prompt Template}
\begin{Verbatim}[breaklines=true,breakanywhere=true,fontsize=\small]
"""
You are a benchmark auditor.

Given description and expected_behavior, evaluate whether expected_behavior is aligned with the description.

Evaluate:
(1) Consistency: expected_behavior must not contradict any explicit facts in description.
(2) No invention: expected_behavior must not introduce new specific scenario facts not supported by description (e.g., names, dates, cities, flight numbers, membership tiers, payment methods, prior actions, constraints).
(3) Coverage: expected_behavior should include all key required outcomes implied by description.
(4) Relevance: expected_behavior should not add unnecessary actions that are not requested or implied by description.

Only judge description and expected_behavior alignment. Do NOT judge writing quality or realism.
Generic statements with no operational meaning are OK.

Return STRICT JSON only.

Score 1–10 for alignment between expected_behavior and description.

Rubric:
- 10: Fully aligned; no invented facts; covers all key user goals; no contradictions.
- 7–9: Mostly aligned; minor missing detail or minor overreach without new specific facts.
- 4–6: Multiple issues: missing key goals OR introduces specific invented facts OR at least one contradiction.
- 1–3: Major mismatch: contradicts description or mostly unrelated behavior.

Return EXACT JSON:
{{
  "score": <1-10>,
  "reasoning": ["3-7 short bullets"]
}}

TASK:
description: {description}
expected_behavior: {expected_behavior}
"""
\end{Verbatim}
\subsection{Policy--Expected Behavior Alignment}

This judge evaluates whether the expected behavior follows the policy constraints for the given scenario.

\paragraph{Inputs}
\begin{itemize}
\item Task description $d_t$
\item Policy specification $P$
\item Expected behavior $e_t$
\end{itemize}

\paragraph{Prompt Template}
\label{fig:Prompt_Template_d_e_p}
\begin{Verbatim}[breaklines=true,breakanywhere=true,fontsize=\small]
"""
You are a benchmark auditor.

Given description, policy, and expected behavior, evaluate whether expected behavior is strictly compliant with policy for the described user request.

Definitions:
- "Act" = any user-facing step or backend step implied by the expected behavior (e.g., asking for identifiers, quoting fees, calling an API / updating a reservation, offering compensation, escalating to a human).
- An act is ALLOWED only if:
  (a) it is explicitly stated in policy, OR
  (b) it is a necessary implication of a policy-required workflow (e.g., policy says "extra baggage is $50" => telling the user "$50" is allowed).
- Any act not allowed by (a) or (b) is an "invented act" (policy violation).
- If expected behavior omits a mandatory step that policy requires for the described workflow, that is also a violation (missing-required-step).
- If expected behavior includes acts not required by description, they must still be permitted by policy; otherwise they are invented acts.

Only judge policy and expected behavior alignment. Do NOT judge writing quality or realism.
Generic statements with no operational meaning (e.g., "be helpful") are OK.

Return STRICT JSON only.

Score 1–10 for alignment between expected behavior and policy.

Rubric:
- 10: Fully compliant; no invented acts; no missing required steps.
- 7–9: Mostly compliant; minor ambiguity or one small missing step that doesn't change allowed actions.
- 4–6: Clear violations (invented acts and/or missing required steps).
- 1–3: Major violations; expected behavior largely contradicts policy or adds many disallowed actions.

Return EXACT JSON:
{{
  "score": <1-10>,
  "reasoning": ["3-7 short bullets"],
}}

TASK:
description: {description}
expected behavior: {expected_behavior}
policy: {policy}
"""
\end{Verbatim}

\subsection{Scoring Procedure}

For each task, the judge produces a single score. Aggregate benchmark scores are computed by averaging task-level scores, while also reporting the median, standard deviation, minimum, and maximum across tasks.

\section{LLM-as-a-Judge Prompt for Policy Violations}
\label{app:judge_violations_prompt}
This judge outputs the IDs of all policy items that are violated by the given task, considering the task description and the initial DB state.

\paragraph{Inputs}

\begin{itemize}
\item Policy specification $P$, annotated with policy item spans $I_P$
\item Task description $d_t$
\item Initial DB state $i_t$
\end{itemize}

\paragraph{Prompt Template}
\begin{Verbatim}[breaklines=true,breakanywhere=true,fontsize=\small]
"""
You are an auditor for conversational agent benchmarks.

The agent guidelines are defined in the following policy document. The document includes annotated spans of specific policy items. Each span is marked as <policy id=...>...</policy>, and has a unique id.

<policy_document>
{policy}
</policy_document>
---

You are given a user scenario that tests the agent's adherence to the above policies. The scenario is a script that guides the user's conversation with the agent. The scenario is visible to the user but not to the agent. 

You are also given relevant rows from the system's database, representing its initial state. 

Your goal is to assess whether any of the user actions in the scenario are not allowed, under the given policy items, and considering the information provided in the initial database state. The policy items should be interpreted in the context of the whole policy document. Output only the policy IDs, not the policies themselves.  

Guidelines:
1. If no user intent is identified in the scenario, you should not output any policy IDs.
2. Do not include policies that are related to the agent's actions rather than to the user's actions, unless the user's actions explicitly violate these policies, or if the user asks the agent to bypass or violate them. 
3. Do not include policies that discuss user authentication, providing information by the user, asking the user to provide information or getting the user's confirmation, unless the scenario explicitly mentions that the user does not provide the required information, or provides incorrect or invalid information.
4. You should output identified violations even if the scenario indicates that they were correctly handled by the agent. 

Response must be a strictly formatted JSON in the following format:
{{
  "policy_ids": [<a list of violated policy IDs, as a list of comma separated integers. Leave empty if no relevant policies were found.>]
}}

Respond with only JSON, no preambles.

TASK:
task_description: {description}
initial_database_state: {db_state}
"""
\end{Verbatim}

\section{Qualitative Examples of Policy--Expected Behavior Alignment Issues}
\label{app:policy_issue_examples}

This appendix presents qualitative examples drawn from the \textbf{Policy--Expected Behavior Alignment} metric. Each example includes a task, its assigned score, and the accompanying judge rationale, illustrating why the task received a low policy-alignment score.

Representative low-scoring policy-alignment examples are shown in
Figures~\ref{fig:policy-gpt4o}, \ref{fig:policy-claude}, and \ref{fig:policy-llama}.

Figure~\ref{fig:policy-gpt4o} illustrates a case with invented workflow stages, forbidden or unsupported actions, and omission of required information collection.
Figure~\ref{fig:policy-claude} shows a different failure mode, where the expected behavior includes a forbidden action and skips explicit confirmation before modification.
Figure~\ref{fig:policy-llama} highlights unsupported operational details, invented workflow constraints, and overly specific refund handling.

Overall, these examples show that the policy-alignment metric is informative not only as a scalar score, but also as a diagnostic signal: low-scoring tasks expose concrete benchmark defects such as unsupported workflow invention, omission of required steps, and direct contradictions of policy rules.

\noindent\textbf{Color legend.}
\textcolor{IssueGreen}{Hallucinated policy stages or non-existent actions},
\textcolor{IssueYellow}{missing required information collection},
\textcolor{IssueCyan}{missing explicit confirmation before action},
\textcolor{IssueMagenta}{incorrect compensation handling},
\textcolor{IssueRed}{forbidden actions}.

\begin{tcolorbox}[
    breakable,
    colback=white,
    colframe=black,
    boxrule=0.6pt,
    arc=2pt,
    left=6pt,right=6pt,top=6pt,bottom=6pt
]
\captionof{figure}{Policy--Expected Behavior misalignment example. 
The low score is driven by an invented modification workflow, a forbidden request for refund method selection, and omission of required cancellation information collection.}
\label{fig:policy-gpt4o}

\textbf{Example 1:
}\\
\textbf{Description.}
User \texttt{john\_doe\_8923} contacts the chatbot regarding their reservation \texttt{RES202310011200}, which includes flight \texttt{F12345}. The user decides to cancel the reservation and inquires about the refund process and cancellation rules.

\smallskip
\textbf{Expected Behavior.}
The chatbot should first assist the user in
\textcolor{IssueGreen}{modifying their flight reservation}
by
\textcolor{IssueRed}{requesting a payment or refund method, such as a gift card or credit card}.
When the user decides to cancel the reservation, the chatbot should inform them that the refund will be processed to the original payment method within 5 to 7 business days. Additionally, the chatbot must ensure that the cancellation rules are strictly followed and communicate these rules to the user before proceeding with the cancellation.

\smallskip
\textbf{Score.} 6

\smallskip
\textbf{Reasoning.}
The \texttt{expected\_behavior}
\textcolor{IssueGreen}{invents an act by assisting the user in modifying their reservation}
and
\textcolor{IssueRed}{requesting a payment or refund method}
before the user has requested a modification; the user only wants to cancel.
\textcolor{IssueYellow}{Policy requires, for cancellation, obtaining user id, reservation id, and reason for cancellation},
then communicating and enforcing cancellation rules before proceeding.
The \texttt{expected\_behavior} correctly states that the refund goes to the original payment method in 5--7 business days and that cancellation rules must be communicated, but the modification/payment method step is not allowed here and is an invented act.
\end{tcolorbox}

\begin{tcolorbox}[
    breakable,
    colback=white,
    colframe=black,
    boxrule=0.6pt,
    arc=2pt,
    left=6pt,right=6pt,top=6pt,bottom=6pt
]
\captionof{figure}{Policy--Expected Behavior misalignment example.
The low score reflects a forbidden post-booking insurance offer and missing explicit confirmation before modifying passenger information.}
\label{fig:policy-claude}
\textbf{Example 2
}\\
\textbf{Description.}
User \texttt{john\_doe\_1980} with ID \texttt{ABC12345} has a reservation \texttt{NYLON123} for a round-trip flight from New York to London (\texttt{NYL123}) and back (\texttt{F2}). The reservation includes passengers \texttt{Alice Smith} (the user) and \texttt{Bob Johnson} (the user's wife). The user requests to update \texttt{Bob Johnson}'s name due to a recent marriage and to add \texttt{P3} (the user's brother) to the reservation. Additionally, the user mentions a previous delay of 3 hours on \texttt{F2} and inquires about refund policies in case of cancellation due to health concerns.

\smallskip
\textbf{Expected Behavior.}
The chatbot should handle this complex request by addressing multiple aspects according to the policies:
(1) The chatbot should acknowledge the user ID (\texttt{ABC12345}) provided and
\textcolor{IssueCyan}{proceed with the modification request}.
(2) The chatbot should inform the user that
\textcolor{IssueCyan}{changing passenger information (wife's name) is possible and can be processed}.
(3) When the user requests adding another passenger (the brother), the chatbot should explain that modifying the number of passengers is not possible; this cannot be done through the system and is not something even a human agent can assist with. The user would need to make a separate booking for the additional passenger.
(4) Regarding the flight delay mentioned by the user, the chatbot should not proactively offer any compensation since the user only mentioned the inconvenience but did not explicitly ask for compensation.
(5) For the question about cancellation and refunds, the chatbot should explain the standard refund policy (refunds go to the original payment method within 5--7 business days) and
\textcolor{IssueRed}{should take this opportunity to offer travel insurance, explaining that it costs \$30 per passenger and would enable a full refund if they need to cancel due to health reasons}.
(6) Throughout the interaction, the chatbot should handle all aspects of the request that fall within its capabilities and should not transfer to a human agent unless specifically requested to handle something outside its scope.

\smallskip
\textbf{Score.} 6

\smallskip
\textbf{Reasoning.}
The \texttt{expected\_behavior} invents an act by
\textcolor{IssueRed}{offering travel insurance after booking, which is explicitly prohibited by policy}
(insurance can only be added at initial booking). It correctly denies adding a passenger, does not offer compensation for the delay, and provides the standard refund policy. However, it
\textcolor{IssueCyan}{misses the mandatory step of obtaining explicit user confirmation before modifying passenger information},
which is required before updating the booking database. These are clear policy violations.

\end{tcolorbox}

\begin{tcolorbox}[
    breakable,
    colback=white,
    colframe=black,
    boxrule=0.6pt,
    arc=2pt,
    left=6pt,right=6pt,top=6pt,bottom=6pt
]
\captionof{figure}{Policy--Expected Behavior misalignment example.
The low score reflects invented workflow constraints, unsupported payment assumptions, and overly specific refund handling.}
\label{fig:policy-llama}
\textbf{Example 3
}\\
\textbf{Description.}
User \texttt{john\_doe\_1234} wants to book flight \texttt{F1} from New York to Los Angeles for passengers \texttt{\{'first\_name': 'Alice', 'last\_name': 'Smith', 'dob': '1985-07-15'\}} and \texttt{\{'first\_name': 'Bob', 'last\_name': 'Johnson', 'dob': '1992-03-22'\}}. The user plans to use payment methods \texttt{\{'source': 'credit\_card', 'brand': 'mastercard', 'last\_four': '5678', 'id': 'credit\_card\_1234567'\}}, \texttt{\{'source': 'certificate', 'amount': 150, 'id': 'certificate\_9876543'\}}, \texttt{\{'source': 'credit\_card', 'brand': 'visa', 'last\_four': '4321', 'id': 'credit\_card\_7654321'\}}, and \texttt{\{'source': 'credit\_card', 'brand': 'amex', 'last\_four': '2223', 'id': 'credit\_card\_1112223'\}} for the payment. The user is interested in purchasing travel insurance. After booking, the user decides to cancel the reservation due to health reasons and requests a refund.

\smallskip
\textbf{Expected Behavior.}
The chatbot should
\textcolor{IssueGreen}{first ask for the user's ID},
then
\textcolor{IssueYellow}{collect the first name, last name, and date of birth for each passenger}.
It should then ask if the user wants to buy travel insurance for \$30 per passenger. The chatbot should inform the user that
\textcolor{IssueGreen}{only one travel certificate, one credit card, and up to three gift cards can be used for the reservation, and that the new gift card needs to be added to the user's profile first}.
\textcolor{IssueGreen}{After the user adds the new gift card},
the chatbot should process the payment and book the flight. When the user wants to cancel the flight, the chatbot should
\textcolor{IssueMagenta}{offer a full refund since the user bought travel insurance}.
The chatbot should not offer any additional compensation unless the user explicitly asks for it.

\smallskip
\textbf{Score.} 5

\smallskip
\textbf{Reasoning.}
The expected behavior
\textcolor{IssueGreen}{introduces several specific constraints and actions not mentioned in the description, such as the limit on payment methods, the need to add a new gift card},
and the travel insurance cost.
It also
\textcolor{IssueGreen}{assumes the user will add a new gift card},
which is not explicitly stated.
\textcolor{IssueMagenta}{The chatbot offering a full refund due to travel insurance is reasonable, but the level of detail is not fully supported by the description.}
\end{tcolorbox}
\section{Benchmark Ordering Scores}
\label{app:ordering_scores}

Table~\ref{tab:metric_ordering_score} reports the Benchmark Ordering Score (\S\ref{ssec:experimental_setup}) for each metric, judge, and domain.

\begin{table*}[t]
\centering
\small
\begin{tabular}{lcccc}
\toprule
 & \multicolumn{3}{c}{Judge} & \\
\cmidrule(lr){2-4}
 & GPT-5.4 & Claude-Sonnet-4.5 & Gemini-2-Flash & Mean \\
\midrule

\multicolumn{5}{l}{\textbf{Airline}} \\
\quad Description--Expected Behavior Alignment & 1.00 & 1.00 & 1.00 & 1.00 \\
\quad Policy--Expected Behavior Alignment    & 1.00 & 1.00 & 1.00 & 1.00 \\
\quad Policy Violations Count                & 1.00 & 1.00 & 1.00 & 1.00 \\
\quad Policy Violations Coverage             & 1.00 & 1.00 & 1.00 & 1.00 \\
\quad \textit{Average}                       & 1.00 & 1.00 & 1.00 & 1.00 \\

\midrule
\multicolumn{5}{l}{\textbf{Retail}} \\
\quad Description--Expected Behavior Alignment & 0.88 & 0.88 & 0.75 & 0.83 \\
\quad Policy--Expected Behavior Alignment    & 1.00 & 0.88 & 1.00 & 0.96 \\
\quad Policy Violations Count                & 1.00 & 1.00 & 1.00 & 1.00 \\
\quad Policy Violations Coverage             & 0.88 & 1.00 & 0.75 & 0.88 \\
\quad \textit{Average}                       & 0.94 & 0.94 & 0.88 & 0.92 \\

\midrule
\textbf{Overall Mean} & 0.97 & 0.97 & 0.94 & 0.96 \\

\bottomrule
\end{tabular}
\caption{\emph{Benchmark Ordering Score} per metric, evaluated across judge models and domains.
\label{tab:metric_ordering_score}}
\end{table*}

\section{Human Validation}
\label{app:human-validation}

\paragraph{Sampling and annotation.}
We sampled 50 tasks from benchmarks produced by all five
benchmark-generator models. A human annotator, one of the
paper's authors, evaluated each task along the same three
task-level measures used by our
framework: Description--Expected Behavior Alignment,
Policy--Expected Behavior Alignment, and Policy Violations Count,
following the same criteria as the corresponding LLM-judge prompts.
The task-level evaluations were pooled across benchmark generators.

\paragraph{Correlation analysis.}
For each task-level measure and LLM judge, we computed Kendall's
$\tau_b$ correlation between the human evaluations and the
corresponding LLM-judge scores. Kendall's $\tau_b$ was selected because
the evaluations contain ordinal scores and ties. We report two-sided
$p$-values. All nine correlations remain statistically significant at
$\alpha=0.05$.

\begin{table}[t]
\centering
\small
\begin{tabular}{llrr}
\toprule
Metric & Judge & \(\tau_b\) & \(p\)-value \\
\midrule
Description--Expected & Claude & 0.46 & \(3.5 \times 10^{-5}\) \\
Description--Expected & GPT & 0.43 & \(8.1 \times 10^{-5}\) \\
Description--Expected & Gemini & 0.35 & \(1.2 \times 10^{-3}\) \\
\midrule
Policy--Expected & Claude & 0.48 & \(7.3 \times 10^{-5}\) \\
Policy--Expected & GPT & 0.55 & \(3.0 \times 10^{-6}\) \\
Policy--Expected & Gemini& 0.32 & \(1.0 \times 10^{-2}\) \\
\midrule
Policy Violations & Claude & 0.56 & \(3.0 \times 10^{-6}\) \\
Policy Violations & GPT  & 0.67 & \(<10^{-6}\) \\
Policy Violations & Gemini  & 0.55 & \(5.0 \times 10^{-6}\) \\
\bottomrule
\end{tabular}
\caption{Task-level Kendall's \(\tau_b\) correlations between human
evaluations and LLM-judge scores. All correlations are positive and
statistically significant.}
\label{tab:human-validation}
\end{table}

Table~\ref{tab:human-validation} reports the results. All correlations
are positive and statistically significant, ranging from
$\tau_b=0.32$ to $\tau_b=0.67$. Agreement is strongest for Policy
Violations Count, for which correlations range from 0.55 to 0.67.
For the two alignment measures, correlations range from 0.32 to
0.55. Overall, these results indicate modest-to-strong rank agreement
between the human evaluations and all three LLM judges across the
three task-level measures.
\section{Evaluation on $\tau$\textsuperscript{3}-Bench}
\label{app:tau3_eval}
This section describes the metric adaptations for evaluating the manually constructed $\tau^3$-Bench. We mapped the \texttt{nl\_assertions} field of $\tau^3$-Bench to our ``expected behavior''. We adapted the metrics as follows:

\begin{itemize}
\item Unlike \textsc{IntellAgent}, which generates an independent, task-relevant initial database state per task, $\tau^3$-Bench tasks rely on a single shared database. We therefore removed the initial DB state $i_t$ from the input to the policy violations judge, and adjusted the prompt accordingly.
\item We revised the policy items to match the updated Airline policy document, using Claude Sonnet 4.6, followed by human review.
\item In \textsc{IntellAgent}, task description and database state are generated together and are expected to contain the same information. In $\tau^3$-Bench, the database and task description are authored separately, so the expected behavior may draw on database information absent from the description. We therefore removed the \emph{``no invention''} criterion from the Description--Expected Behavior metric (App.~\ref{app:desc_expected}).
\end{itemize}

\section{Usage of AI Assistants}
We have used Claude Sonnet 5 as a writing assistant, in accordance with ACL/ARR guidelines.

\end{document}